\documentclass{article}
\usepackage{ijcai11}
\usepackage{times}

\usepackage{algorithm2e}
\usepackage[dvips]{epsfig}
\usepackage{graphicx}
\usepackage{times}
\usepackage{colortbl}
\usepackage{multirow}
\usepackage{url}

\title{{\bf SPPAM - Statistical PreProcessing AlgorithM \\ An approach
    for the classification of multiple correlated data}}
\author{{Tiago Silva, In\^es Dutra}\\
CRACS \& INESC-Porto LA \\
Faculdade de Ci\^encias, Universidade do Porto  \\
Rua do Campo Alegre, 1021 \\
4169-007 Porto PORTUGAL \\
{\it tiagosilva@inbox.com, ines@dcc.fc.up.pt}
}

\begin{document}
\maketitle

\begin{abstract}
Most machine learning tools work with a single table where each row is
an instance and each column is an attribute. Each cell of the table
contains an attribute value for an instance. This representation
prevents one important form of learning, which is, classification
based on groups of correlated records, such as multiple exams of a
single patient, internet customer preferences, weather forecast or
prediction of sea conditions for a given day. To some extent,
relational learning methods, such as inductive logic programming, can
capture this correlation through the use of intensional predicates
added to the background knowledge. In this work, we propose SPPAM, an
algorithm that aggregates past observations in one single record. We
show that applying SPPAM to the original correlated data, before the
learning task, can produce classifiers that are better than the ones
trained using all records.
\end{abstract}
{\bf Keywords:} multi-relational data,
classification, data preprocessing.

\section{Introduction}

Machine learning techniques have been successfully applied to various
domains. However there is a lack of formal methodology and application
of machine learning tools to datasets that are characterized by
subgroups of correlated records. Examples are medical records with
multiple exams of a single patient, internet customer preferences,
weather forecast and prediction of sea conditions, among others. Despite
the fact that there are many applications that fall into this
category, there is also a lack of available datasets with this
characteristic in the main UCI machine learning repository
(\url{http://archive.ics.uci.edu/ml/}).

Machine learning tools usually learn classifiers from a single table
where each row is an instance and each column is an attribute. Each
cell of the table contains an attribute value for an instance. Most of
these tools treat each row of this table as independent from each
other, which prevents one important form of learning based on groups
of correlated records. Tools based on first order logic (inductive
logic programming) can partially overcome this problem because they
can do multi-relational learning. But first order rules in the form of
intensional predicates need to be added to the background knowledge,
in order to code the multi-relational meaning intended by the
observer~\cite{amia05}.

When dealing with data that have this multi-relational characteristic,
one additional problem arises when using cross-validation. Ideally,
records that belong to the same observation period need to be manually
separated in a way that all records of a certain period falls into
just one fold. Machine learning tools like WEKA~\cite{weka}, for
example, do not allow training based on pre-defined folds (unless when
using the percentage split training option).

In this work, we propose a general method that connects records that
are correlated (either through the same location or observation
period). To the best of our knowledge, this is the first work that
tackles this problem.

We propose SPPAM (Statistical PreProcessing AlgorithM), an algorithm
that aggregates past correlated observations in one single record. We
apply SPPAM to two datasets of surf conditions. Our task is to learn a
classifier that predicts well if a certain beach is adequate for
surfing in a certain day. We perform our experiments using the WEKA
machine learning tool and compare the performance of various WEKA
algorithms trained on the original datasets and on the
SPPAM-transformed datasets. We show that applying SPPAM to the
original correlated data, before the learning task, can produce
classifiers that are better than the ones trained using the original
datasets with all records.

Some work has been done on characterizing
relations from weather observations and forecasts using machine
learning techniques. For example, Ingsrisawan {\it et al.} used
support vector machines, decision trees and neural networks to develop
models to predict rainfall occurrences in
Thailand~\cite{ingsrisawang08}. Lai {\it et al.} proposed a
preprocessing technique for weather data in order to predict
temperature and weather conditions~\cite{Lai04}. Williams{\it et al.}
proposed the use of Random Forests to predict and classify storm
forecastings~\cite{williams08}. However, we are not only interested in
temporal patterns nor pure weather forecast. Our goal is to provide a
generic preprocessing technique that increase the classification task's
performance for every suitable dataset.


This paper is organized as follows.
In Section~\ref{alg} we describe the SPPAM algorithm. In
Section~\ref{met} we discuss the methodology used to run our
experiments and present the datasets. In Section~\ref{res1}, we
present and discuss our results. Finally, in Section~\ref{conc} we
draw our conclusions and give perspectives of future work.

\section{An approach to classify 
  multiple correlated data}
\label{alg}

SPPAM is a two-step algorithm that captures the hierarchical aspect of
learning from a dataset with multiple records for the same observation
period (or location). The first step is to separate and consolidate
records that belong to the same time/location interval. The user needs
to provide the name of the attribute that will be used to perform this
separation and the name of the class attribute. We also assume that
each record has a unique identifier. We use a transformation that maps
several records into just one record along with a transformation on
the original attributes. The algorithm can be seen in
Algorithm~\ref{alg:alg1}.

\begin{algorithm}[htbp]
\label{alg:alg1}
\SetLine
\KwData{\\ $Dataset$, \emph{// original dataset}}
\KwResult{\\ $Out$, \emph{// original dataset transformed}}
\BlankLine
Initialize a new empty dataset $Out$\;
Read $Dataset$\;
\ForEach{Attribute $a$ in $Dataset$}{
\If{ $a$ has type Numeric}{
    create Attributes $a$-Maximum, $a$-Minimum, $a$-Average and $a$-Last on $Out$\;
}
\ElseIf{ $a$ is Nominal}{
    create Attributes $a$-Frequency for each nominal value and $a$-Last on $Out$\;
}
\Else{
    copy $a$ to $Out$\;
}
}
Group correlated records according to the user provided field\;
\ForEach{Group $i$}{
read each individual attribute value A\;
\If{ A has type String or is the ID}{
  copy A to $Out$\;
}
\If{ A has type numeric }{
  calculate Maximum, Minimum, Average and Last values among all values of A
  for group $i$\;
  copy them to $Out$\;
}
\If{ A is nominal}{
  copy frequency and the last value of A in group $i$ to $Out$\;
}
Take the value of the class variable of last instance of Group $i$\;
Copy it to $Out$ to complete the record
}
\caption{The SPPAM algorithm}
\end{algorithm}

This basic version of the algorithm maps groups of records of each
observation to just one record by computing aggregates for the values
of the attributes. But what to do with the class variable? In this
algorithm, we keep the last class value of the group (i.e., the most
recent observation).

Figures~\ref{fig:ex1} and~\ref{fig:ex2} illustrate an example of this
transformation.

\begin{figure}[htbp]
\centering
\begin{center}
{\tiny
\begin{verbatim}
@ATTRIBUTE Date String
@ATTRIBUTE Wind_Knots numeric
@ATTRIBUTE Wind_Dir {N, NE, E, SE, S, SW, W, NW}
@ATTRIBUTE Surf {0,1}
@DATA
18-11-2010,15.6,SE,0
18-11-2010,9.7,SE,0
18-11-2010,3.9,SE,0
18-11-2010,5.8,NE,0
19-11-2010,11.7,NE,0
19-11-2010,15.6,NE,0
19-11-2010,13.6,E,1
19-11-2010,15.6,E,1
\end{verbatim}
}
\end{center}
\caption{Original dataset}
\label{fig:ex1}
\end{figure}

\begin{figure}[htbp]
\centering
{\tiny
\begin{verbatim}
@ATTRIBUTE Date STRING
@ATTRIBUTE Wind_Knots_MAX NUMERIC
@ATTRIBUTE Wind_Knots_MIN NUMERIC
@ATTRIBUTE Wind_Knots_AVG NUMERIC
@ATTRIBUTE Wind_Knots_LAST NUMERIC
@ATTRIBUTE Wind_Dir_N_PERC NUMERIC
@ATTRIBUTE Wind_Dir_NE_PERC NUMERIC
@ATTRIBUTE Wind_Dir_E_PERC NUMERIC
@ATTRIBUTE Wind_Dir_SE_PERC NUMERIC
@ATTRIBUTE Wind_Dir_S_PERC NUMERIC
@ATTRIBUTE Wind_Dir_SW_PERC NUMERIC
@ATTRIBUTE Wind_Dir_W_PERC NUMERIC
@ATTRIBUTE Wind_Dir_NW_PERC NUMERIC
@ATTRIBUTE Wind_Dir_LAST {N, NE, E, SE, S, SW, W, NW}
@ATTRIBUTE Surf {0,1}
@DATA
18-11-2010,15.6,3.9,8.75,5.8,0.0,25.0,0.0,75.0,0.0,0.0,0.0,0.0,NE,0
19-11-2010,15.6,11.7,14.13,15.6,0.0,50.0,50.0,0.0,0.0,0.0,0.0,0.0,E,1
\end{verbatim}
}
\caption{SPPAM-transformed dataset}
\label{fig:ex2}
\end{figure}

The original dataset (Figure~\ref{fig:ex1}) shows an example of two
days of observation of weather and sea conditions for surf practice
with 4 attributes and 8 instances. The first attribute \emph{Date} is
of type String and will be our aggregation pivot attribute, the second
is of type numeric, the third attribute is nominal (with eight
possible values) and the last attribute (the class) is binary. Our
goal is to aggregate all observations within a day in one single
record.

The transformed dataset for this example has 2 data rows for two
observation days (18-11-2010 and 19-11-2010), each with 15
attributes. The first attribute is the date. The next 4 attributes are
numeric values corresponding to the maximum, minimum, average and last
values of the \emph{Wind\_Knots} attribute. The following 8 numeric
values correspond to the frequencies of each nominal value of the
attribute \emph{Wind\_Dir}. The following attribute (14) is a nominal
value representing the last observed value for the \emph{Wind\_Dir}
attribute. The last attribute is the last value of the class attribute for the group. The number
of instances of the transformed dataset drops to only 2 given that we
had only 2 complete days of observation in the original table. For
attribute 2, \emph{Wind\_Knots}, the first day has maximum value of
15.6, minimum 3.9 and average of 8.75 and the last obtained value
5.8. The same would repeat for another hypothetical numeric
attribute. Attribute \emph{Wind\_Dir} in the original dataset will
unfold in nine attributes on the transformed data, because it is
nominal and it has eight values plus the last observed value. The
first unfolded value corresponds to the frequency of occurrence of the
first value on that observation group. The second, to the frequency of
occurrence of the second value and so on.

The total number of attributes on the transformed dataset is given by
the equation
\begin{equation}
1 + s + 4n + \sum_{i=1}^{w}\left ( V\left ( w{_{i}} \right ) + 1 \right )
\label{eq:attrs}
\end{equation}
where \emph{s} is the number of String attributes on the original
dataset, \emph{n} is the number of numeric attributes on the original
dataset, \emph{w} is the number of nominal attributes and the function
\emph{V(w)} is the number of values of the nominal attribute \emph{w},
for all non-class attributes.

The number of records on the transformed
dataset is equal to the number of different unique \emph{ids} on the
original datasets, in our example the \emph{id} is the date attribute.

After this preprocessing task, the second step is to feed the new
table (transformed dataset) to a machine learning algorithm, like any
other dataset.

Although we are dealing with meteorological data, the method above
described is fully applicable to any kind of relational data where
various records are related to the same individual.

\section{Methodology and Applications}
\label{met}
We applied our algorithm to two datasets. The first one is the Surf -
Praia Grande dataset which has 10 attributes, 5 of them numeric, 4
nominal and 1 string. This dataset contains four daily observations of
wind and sea conditions taken from the Praia Grande beach, Portugal,
between November 18th 2010 and January 6th 2011, in the total of 192
instances. The 10 attributes are: date, hour, total sea height, wave
height, wave direction, wind wave height, wind speed, wind direction,
water temperature and wave set quality to practice surf. This last
attribute is our class which can have 2 different values: 0 and 1,
where 0 means that the weather and sea conditions are not good for surf
practice, and 1 means that there are good conditions to surf.

The second dataset is the Surf - Aljezur and it has the same structure
(data were collected at the same period of time as Praia Grande). The
attributes and number of instances are the same as the Surf - Praia
Grande.

Table~\ref{tab:original_dataset} shows the detailed structure of the
original datasets to be transformed by the SPPAM algorithm.
A summary of the transformations on both datasets is shown in
table~\ref{tab:sppam_transformations}.

For both datasets, the number of attributes generated by SPPAM is 44
(which follows from equation~\ref{eq:attrs}) and the number of
instances is 48 (the number of different observation days).

\begin{table}[htb]
\label{tab:original_dataset}
\begin{center}
\caption{Original Surf - Praia Grande and Original Surf - Aljezur attributes}
{\tiny
\begin{tabular}{|l|l|l|} \hline
\textbf{Attribute} & \textbf{Type} & \textbf{Values}\\
\hline
Date & String & \\
\hline
Hour & Nominal & 0, 6, 12, 18 \\
\hline
Wave Total & Numeric & \\
\hline
Wave & Numeric & \\
\hline
Wave Direction & Nominal & N, NE, E, SE, S, SW, W, NW \\
\hline
Vaga & Numeric & \\
\hline
Wind Speed & Numeric & \\
\hline
Wind Direction & Nominal & N, NE, E, SE, S, SW, W, NW \\
\hline
Water Temperature & Numeric & \\
\hline
Sets & Nominal (Class) & 0,1 \\
\hline
\end{tabular}
}
\end{center}
\end{table}

\begin{table}
\begin{center}
\caption{Original and SPPAM-transformed datasets summary}
{\tiny
\begin{tabular}{|l|l|l|l|}
\hline
\textbf{Dataset} & \textbf{\# Instances} & \textbf{Class = 0}  & \textbf{Class = 1} \\
\hline
Sintra & 192 & 75 (39\%) & 117 (61\%)\\
\hline
Sintra SPPAM & 48 & 18 (38\%) & 30 (62\%)\\
\hline
\hline
Aljezur & 192 & 48 (25\%) & 144 (75\%)\\
\hline
Aljezur SPPAM & 48 & 9 (19\%) & 39 (81\%)\\
\hline
\end{tabular}
}
\label{tab:sppam_transformations}
\end{center}

\end{table}

After applying our algorithm to the datasets, we performed learning
experiments using the WEKA tool, developed at Waikato University, New
Zealand~\cite{weka}.
The experiments were performed in WEKA using the Experimenter
module, where we set several parameters, including the statistical
significance test and confidence interval, and the algorithms we
wanted to use (we used OneR as reference, ZeroR, PART, J48,
SimpleCart, DecisionStump, Random Forests, SMO, Naive Bayes, Bayes
with TAN, NBTree and DTNB). The WEKA experimenter produces a table
with the performance metrics of all algorithms with an indication of
statistical differences, using one of the algorithms as a
reference. The significance tests were performed using standard
corrected t-test with a significance level of 0.01. The parameters
used for the learning algorithms are the WEKA defaults. For all
experiments we used 10-fold stratified cross-validation and report
results for the test sets.

\section{Results}
\label{res1}

We compared the results obtained in WEKA using our preprocessing
method SPPAM with the results obtained with the original datasets. In
tables~\ref{tab:results1} and ~\ref{tab:results2}, we present the
performance obtained by the WEKA algorithms for both the original
dataset and the SPPAM transformed dataset for Surf - Praia Grande and
Surf - Aljezur. We show the results obtained for Percentage of
Correctly Classified Instances (CCI), Kappa Statistic (Kappa),
Precision (Precis.), Recall and F-Measure (F-Meas.). We show the
performance for each class and the averaged performance for both
classes. Our best results with SPPAM are highlighted on both tables. We
also present charts showing the average performance gain between the
correctly classified instances average for the SPPAM datasets
and for the original datasets on all classification algorithms.

\subsection{Praia Grande dataset results}
For this particular dataset, our best results were obtained using
Bayesian Networks (using the TAN and K2 search algorithms), Naive
Bayes and DTNB, as shown in Table~\ref{tab:results1}. Naive Bayes is
the algorithm that yields the best performance when training with the
SPPAM-transformed datasets, for every metric.

\begin{table}[Hhtb]
\begin{center}
\caption{Transformed Surf - Praia Grande results}
{\tiny
\begin{tabular}{|@{}l@{}|@{}r@{}|@{}c@{}|@{}c@{}|@{}c@{}|@{}c@{}|@{}c@{}||@{}c@{}|@{}c@{}|@{}c@{}|@{}c@{}|@{}c@{}|}
\cline{3-12}
\multicolumn{2}{c|}{} & \multicolumn{5}{c||}{\textbf{Original Dataset}} & \multicolumn{5}{c|}{\textbf{SPPAM transformed dataset}}\\
\cline{3-12}
\multicolumn{2}{c|}{} & \textbf{CCI\%} & \textbf{Kappa} & \textbf{Precis.} & \textbf{Recall} & \textbf{F-Meas.} & \textbf{CCI\%} & \textbf{Kappa} & \textbf{Precis.} & \textbf{Recall} & \textbf{F-Meas.}\\
\cline{3-12}
\hline
\multirow{3}{*}{\textbf{Bayes Net(K2)}} & \emph{Class=0} & 73.44 & 0.45 & 0.67 & 0.68 & 0.66 & \textbf{76.75} & \textbf{0.50} & 0.66 & \textbf{0.73} & 0.66\\
\cline{2-12}
 & \emph{Class=1} & 72.92 & 0.44 & 0.80 & 0.75 & 0.77 & \textbf{79.10} & \textbf{0.53} & \textbf{0.86} & \textbf{0.83} & \textbf{0.83}\\
\cline{2-12}
 & \emph{Average} & 73.18 & 0.45 & 0.74 & 0.72 & 0.72 & \textbf{77.93} & \textbf{0.52} & \textbf{0.76} & \textbf{0.78} & \textbf{0.75}\\
\hline
\multirow{3}{*}{\textbf{Bayes Net(TAN)}} & \emph{Class=0} & 74.64 & 0.46 & 0.69 & 0.66 & 0.66 & \textbf{76.70} & \textbf{0.50} & \textbf{0.68} & \textbf{0.72} & \textbf{0.67}\\
\cline{2-12}
 & \emph{Class=1} & 75.42 & 0.48 & 0.80 & 0.81 & 0.80 & \textbf{79.65} & \textbf{0.55} & \textbf{0.88} & 0.81 & \textbf{0.82}\\
\cline{2-12}
 & \emph{Average }& 75.03 & 0.47 & 0.75 & 0.74 & 0.73 & \textbf{78.18} & \textbf{0.53} & \textbf{0.78} & \textbf{0.77} & \textbf{0.75}\\
\hline
\multirow{3}{*}{\textbf{Naive Bayes}} & \emph{Class=0} & 70.43 & 0.41 & 0.61 & 0.75 & 0.66 & \textbf{77.20} & \textbf{0.54} & \textbf{0.66} & \textbf{0.84} & \textbf{0.72}\\
\cline{2-12}
 & \emph{Class=1} & 71.04 & 0.42 & 0.82 & 0.68 & 0.74 & \textbf{80.00} & \textbf{0.58} & \textbf{0.94} & \textbf{0.76} & \textbf{0.81}\\
\cline{2-12}
 & \emph{Average} & 70.74 & 0.42 & 0.72 & 0.72 & 0.70 & \textbf{78.60} & \textbf{0.56} & \textbf{0.80} & \textbf{0.80} & \textbf{0.77}\\
\hline
\multirow{3}{*}{\textbf{SMO} } & \emph{Class=0} & 75.80 & 0.50 & 0.70 & 0.71 & 0.69 & 75.60 & 0.48 & 0.69 & 0.71 & 0.66\\
\cline{2-12}
 & \emph{Class=1} & 76.54 & 0.51 & 0.82 & 0.81 & 0.81 & 75.60 & 0.45 & \textbf{0.83} & 0.81 & 0.80\\
\cline{2-12}
 & \emph{Average} & 76.17 & 0.51 & 0.76 & 0.76 & 0.75 & 75.60 & 0.47 & 0.76 & 0.76 & 0.73\\
\hline
\multirow{3}{*}{\textbf{Decision Stump}} & \emph{Class=0} & 71.68 & 0.42 & 0.63 & 0.69 & 0.65 & 67.00 & 0.30 & 0.51 & 0.60 & 0.52\\
\cline{2-12}
 & \emph{Class=1}& 72.76 & 0.44 & 0.81 & 0.73 & 0.76 & 72.15 & 0.39 & \textbf{0.83} & \textbf{0.76} & 0.77\\
\cline{2-12}
 & \emph{Average} & 72.22 & 0.43 & 0.72 & 0.71 & 0.71 & 69.58 & 0.35 & 0.67 & 0.68 & 0.65\\
\hline
\multirow{3}{*}{\textbf{J48} } & \emph{Class=0} & 77.66 & 0.52 & 0.79 & 0.63 & 0.68 & 65.75 & 0.24 & 0.51 & 0.50 & 0.47\\
\cline{2-12}
 & \emph{Class=1} & 77.26 & 0.51 & 0.79 & 0.86 & 0.82 & 64.55 & 0.20 & 0.75 & 0.73 & 0.71\\
\cline{2-12}
 & \emph{Average} & 77.46 & 0.52 & 0.79 & 0.75 & 0.75 & 65.15 & 0.22 & 0.63 & 0.62 & 0.59\\
\hline
\multirow{3}{*}{\textbf{NB Tree}} & \emph{Class=0} & 76.88 & 0.50 & 0.75 & 0.64 & 0.68 & 71.60 & 0.40 & 0.57 &\textbf{ 0.68} & 0.60\\
\cline{2-12}
 & \emph{Class=1} & 76.52 & 0.49 & 0.79 & 0.84 & 0.81 & 76.50 & 0.48 & \textbf{0.84} & 0.78 & 0.79\\
\cline{2-12}
 & \emph{Average} & 76.70 & 0.50 & 0.77 & 0.74 & 0.75 & 74.05 & 0.44 & 0.71 & 0.73 & 0.70\\
\hline
\multirow{3}{*}{\textbf{Random Forest} } & \emph{Class=0} & 78.29 & 0.55 & 0.73 & 0.74 & 0.73 & 70.55 & 0.38 & 0.61 & 0.67 & 0.60\\
\cline{2-12}
 & \emph{Class=1} & 78.76 & 0.55 & 0.83 & 0.84 & 0.83 & 75.90 & 0.43 & 0.81 &\textbf{ 0.85} & 0.82\\
\cline{2-12}
 & \emph{Average} & 78.53 & 0.55 & 0.78 & 0.79 & 0.78 & 73.23 & 0.41 & 0.71 & 0.76 & 0.71\\
\hline
\multirow{3}{*}{\textbf{Simple CART}} & \emph{Class=0} & 74.71 & 0.46 & 0.71 & 0.65 & 0.67 & 66.80 & 0.29 & 0.51 & 0.61 & 0.52\\
\cline{2-12}
 & \emph{Class=1} & 74.40 & 0.46 & 0.79 & 0.81 & 0.79 & 71.75 & 0.37 & \textbf{0.82} & 0.76 & 0.77\\
\cline{2-12}
 & \emph{Average} & 74.56 & 0.46 & 0.75 & 0.73 & 0.73 & 69.28 & 0.33 & 0.67 & 0.69 & 0.65\\
\hline
\multirow{3}{*}{\textbf{DTNB} } & \emph{Class=0} & 74.14 & 0.45 & 0.69 & 0.66 & 0.66 & \textbf{75.25} &\textbf{ 0.47} & 0.67 & \textbf{0.69} & 0.65\\
\cline{2-12}
 & \emph{Class=1} & 74.85 & 0.46 & 0.79 & 0.81 & 0.80 & \textbf{76.45} & \textbf{0.47} & \textbf{0.83} & \textbf{0.82} & \textbf{0.81}\\
\cline{2-12}
 & \emph{Average} & 74.50 & 0.46 & 0.74 & 0.74 & 0.73 & \textbf{75.85} & \textbf{0.47} & \textbf{0.75} & \textbf{0.76} & 0.73\\
\hline
\multirow{3}{*}{\textbf{PART} } & \emph{Class=0} & 76.92 & 0.51 & 0.72 & 0.70 & 0.70 & 67.10 & 0.28 & 0.54 & 0.55 & 0.51\\
\cline{2-12}
 & \emph{Class=1} & 75.86 & 0.49 & 0.80 & 0.82 & 0.80 & 66.20 & 0.25 & 0.77 & 0.73 & 0.72\\
\cline{2-12}
 & \emph{Average} & 76.39 & 0.50 & 0.76 & 0.76 & 0.75 & 66.65 & 0.27 & 0.66 & 0.64 & 0.62\\
\hline
\multirow{3}{*}{\textbf{ZeroR}} & \emph{Class=0} & 60.95 & 0.00 & 0.00 & 0.00 & 0.00 &\textbf{ 63.00} & 0.00 & 0.00 & 0.00 & 0.00\\
\cline{2-12}
 & \emph{Class=1} & 61.47 & 0.00 & 0.61 & 1.00 & 0.76 &\textbf{ 65.00} & 0.00 & \textbf{0.65} & 1.00 & \textbf{0.79}\\
\cline{2-12}
 & \emph{Average} & 61.21 & 0.00 & 0.31 & 0.50 & 0.38 & \textbf{64.00} & 0.00 & \textbf{0.33} & 0.50 & \textbf{0.40}\\
\hline
\multirow{3}{*}{\textbf{OneR}} & \emph{Class=0} & 70.41 & 0.37 & 0.65 & 0.60 & 0.60 & 70.25 & 0.33 & 0.54 & 0.56 & 0.52\\
\cline{2-12}
 & \emph{Class=1} & 72.07 & 0.41 & 0.77 & 0.79 & 0.77 & 70.95 & 0.33 & \textbf{0.79} & 0.79 & 0.77\\
\cline{2-12}
 & \emph{Average} & 71.24 & 0.39 & 0.71 & 0.70 & 0.69 & 70.60 & 0.33 & 0.67 & 0.68 & 0.65\\
\hline
\end{tabular}
}
\label{tab:results1}
\end{center}
\end{table}

\begin{figure}
  \includegraphics[width=3.5in]{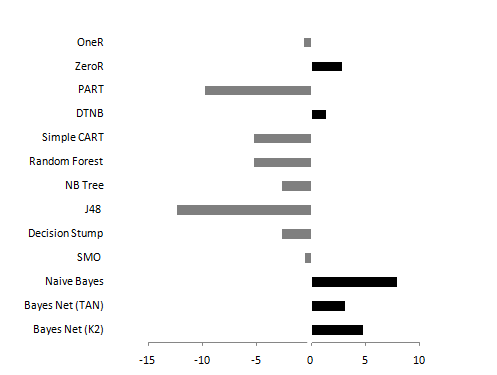}
  \caption{Delta between average CCI\% Praia Grande and Praia Grande SPPAM }
  \label{fig:sintra_delta}
\end{figure}

In Figure~\ref{fig:sintra_delta} we show graphically the differences
between the correctly classified instances percentage average for the
original dataset and the SPPAM-preprocessed dataset for the Praia
Grande data for all the machine learning algorithms we tested. The
values are in percentage.

\subsection{Aljezur dataset results}
\label{res2}
With this dataset we achieved even better results. The use of SPPAM before
the training task improved the classification performance on almost
all analyzed metrics. In some cases, we get 10\% gain on the correctly
classified instances percentage. These results were statistical
significant for BayesNet using K2 and TAN, SimpleCart, ZeroR, SMO and
DTNB. For the metrics where the results were not improved (Naive
Bayes and DTNB), the difference is not significant.

\begin{table}[Hhtb]
\begin{center}
\caption{Transformed Surf - Aljezur results}
{\tiny
\begin{tabular}{|@{}l@{}|@{}r@{}|@{}c@{}|@{}c@{}|@{}c@{}|@{}c@{}|@{}c@{}||@{}c@{}|@{}c@{}|@{}c@{}|@{}c@{}|@{}c@{}|}
\cline{3-12}
\multicolumn{2}{c|}{} & \multicolumn{5}{c||}{\textbf{Original Dataset}} & \multicolumn{5}{c|}{\textbf{SPPAM transformed dataset}}\\
\cline{3-12}
\multicolumn{2}{c|}{} & \textbf{CCI\%} & \textbf{Kappa} & \textbf{Precis.} & \textbf{Recall} & \textbf{F-Meas.} & \textbf{CCI\%} & \textbf{Kappa} & \textbf{Precis.} & \textbf{Recall} & \textbf{F-Meas.}\\
\cline{3-12}
\hline
\multirow{3}{*}{\textbf{Bayes Net(K2)}} & \emph{Class=0} & 73.84 & -0.02 & 0.00 & 0.00 & 0.00 & \textbf{81.05} & \textbf{0.09} & 0.00 & 0.00 & 0.00\\
\cline{2-12}
 & \emph{Class=1} & 74.00 & -0.02 & 0.75 & 0.98 & 0.85 & \textbf{83.25} & \textbf{0.19} & \textbf{0.84} & \textbf{0.99} & \textbf{0.91}\\
\cline{2-12}
 & \emph{Average} & 73.92 & -0.02 & 0.38 & 0.49 & 0.43 & \textbf{82.15} & \textbf{0.14} & \textbf{0.42} & \textbf{0.50} & \textbf{0.46}\\
\hline
\multirow{3}{*}{\textbf{Bayes Net(TAN)}} & \emph{Class=0} & 72.75 & -0.02 & 0.04 & 0.02 & 0.03 & \textbf{80.65} & \textbf{0.08} & 0.00 & 0.00 & 0.00\\
\cline{2-12}
 & \emph{Class=1} & 73.07 & -0.03 & 0.75 & 0.96 & 0.84 & \textbf{83.25} & \textbf{0.19} & \textbf{0.84} & \textbf{0.99} & \textbf{0.91}\\
\cline{2-12}
 & \emph{Average} & 72.91 & -0.03 & 0.40 & 0.49 & 0.44 & \textbf{81.95} & \textbf{0.14} & \textbf{0.42} & \textbf{0.50} & \textbf{0.46}\\
\hline
\multirow{3}{*}{\textbf{Naive Bayes}} & \emph{Class=0} & 64.10 & 0.04 & 0.27 & 0.28 & 0.27 & 62.55 & -0.01 & 0.12 & 0.22 & 0.15\\
\cline{2-12}
 & \emph{Class=1} & 63.52 & 0.05 & 0.77 & 0.74 & 0.75 & 61.95 & -0.05 & \textbf{0.82} & 0.71 & 0.74\\
\cline{2-12}
 & \emph{Average} & 63.81 & 0.05 & 0.52 & 0.51 & 0.51 & 62.25 & -0.03 & 0.47 & 0.47 & 0.45\\
\hline
\multirow{3}{*}{\textbf{SMO} } & \emph{Class=0} & 75.00 & 0.00 & 0.00 & 0.00 & 0.00 & \textbf{79.80} & \textbf{0.07} & 0.00 & 0.00 & 0.00\\
\cline{2-12}
 & \emph{Class=1 }& 75.53 & 0.00 & 0.76 & 1.00 & 0.86 & \textbf{80.45} & \textbf{0.10} &\textbf{0.84} & 0.96 & \textbf{0.89}\\
\cline{2-12}
 & \emph{Average} & 75.27 & 0.00 & 0.38 & 0.50 & 0.43 & \textbf{80.13} & \textbf{0.09} & \textbf{0.42} & 0.48 & \textbf{0.45}\\
\hline
\multirow{3}{*}{\textbf{Decision Stump}} & \emph{Class=0} & 73.90 & 0.00 & 0.06 & 0.03 & 0.03 & \textbf{80.20} & \textbf{0.07} & 0.00 & 0.00 & 0.00\\
\cline{2-12}
 & \emph{Class=1} & 75.16 & 0.05 & 0.76 & 0.98 & 0.86 & \textbf{82.75} & \textbf{0.18} & \textbf{0.84} & \textbf{0.99} & \textbf{0.90}\\
\cline{2-12}
 & \emph{Average} & 74.53 & 0.03 & 0.41 & 0.51 & 0.45 & \textbf{81.48} & \textbf{0.13} & \textbf{0.42} & 0.50 & 0.45\\
\hline
\multirow{3}{*}{\textbf{J48} } & \emph{Class=0} & 73.27 & 0.02 & 0.12 & 0.06 & 0.07 & \textbf{74.50} & -0.02 & 0.00 & 0.00 & 0.00\\
\cline{2-12}
 & \emph{Class=1} & 74.37 & 0.03 & 0.76 & 0.97 & 0.85 & \textbf{77.60} & \textbf{0.10} & \textbf{0.83} & 0.93 & \textbf{0.87}\\
\cline{2-12}
 & \emph{Average} & 73.82 & 0.03 & 0.44 & 0.52 & 0.46 & \textbf{76.05} & \textbf{0.04} & 0.42 & 0.47 & 0.44\\
\hline
\multirow{3}{*}{\textbf{NB Tree}} & \emph{Class=0} & 73.74 & 0.04 & 0.17 & 0.07 & 0.09 & \textbf{79.20} & \textbf{0.08} & 0.00 & 0.00 & 0.00\\
\cline{2-12}
 & \emph{Class=1} & 74.32 & 0.05 & 0.76 & 0.96 & 0.85 & \textbf{80.35} & \textbf{0.16} & \textbf{0.83} & 0.95 & \textbf{0.88}\\
\cline{2-12}
 & \emph{Average} & 74.03 & 0.05 & 0.47 & 0.52 & 0.47 & \textbf{79.78} & \textbf{0.12} & 0.42 & 0.48 & 0.44\\
\hline
\multirow{3}{*}{\textbf{Random Forest }} & \emph{Class=0} & 69.16 & 0.19 & 0.40 & 0.41 & 0.39 & \textbf{70.70} & -0.05 & 0.01 & 0.02 & 0.01\\
\cline{2-12}
 & \emph{Class=1} & 72.55 & 0.20 & 0.81 & 0.84 & 0.82 & \textbf{79.05} & 0.15 & \textbf{0.82 }& 0.94 & \textbf{0.87}\\
\cline{2-12}
 & \emph{Average} & 70.86 & 0.20 & 0.61 & 0.63 & 0.61 & \textbf{74.88} & 0.05 & 0.42 & 0.48 & 0.44\\
\hline
\multirow{3}{*}{\textbf{Simple CART}} & \emph{Class=0} & 73.28 & -0.01 & 0.04 & 0.03 & 0.03 & \textbf{81.50} & \textbf{0.10} & 0.00 & 0.00 & 0.00\\
\cline{2-12}
 & \emph{Class=1} & 74.63 & 0.01 & 0.76 & 0.98 & 0.85 & \textbf{84.00} & \textbf{0.20} & \textbf{0.84} & \textbf{1.00} & \textbf{0.91}\\
\cline{2-12}
 & \emph{Average} & 73.96 & 0.00 & 0.40 & 0.51 & 0.44 & \textbf{82.75} & \textbf{0.15} & \textbf{0.42} & 0.50 & \textbf{0.46}\\
\hline
\multirow{3}{*}{\textbf{DTNB} } & \emph{Class=0} & 72.70 & -0.02 & 0.03 & 0.02 & 0.02 & \textbf{77.95} & \textbf{0.05} & 0.00 & 0.00 & 0.00\\
\cline{2-12}
 & \emph{Class=1} & 73.54 & -0.02 & 0.75 & 0.97 & 0.85 & \textbf{81.00} & \textbf{0.14} & \textbf{0.83} & 0.97 & \textbf{0.89}\\
\cline{2-12}
 & \emph{Average} & 73.12 & -0.02 & 0.39 & 0.50 & 0.44 & \textbf{79.48} & \textbf{0.10} & \textbf{0.42} & 0.49 & \textbf{0.45}\\
\hline
\multirow{3}{*}{\textbf{PART} } & \emph{Class=0} & 67.23 & 0.09 & 0.34 & 0.29 & 0.29 & 66.10 & -0.10 & 0.02 & 0.04 & 0.02\\
\cline{2-12}
 & \emph{Class=1} & 67.52 & 0.05 & 0.77 & 0.82 & 0.79 & \textbf{68.95} & -0.04 & \textbf{0.80} & 0.82 & \textbf{0.80}\\
\cline{2-12}
 & \emph{Average} & 67.38 & 0.07 & 0.56 & 0.56 & 0.54 & \textbf{67.53} & -0.07 & 0.41 & 0.43 & 0.41\\
\hline
\multirow{3}{*}{\textbf{ZeroR}} & \emph{Class=0} & 75.00 & 0.00 & 0.00 & 0.00 & 0.00 & \textbf{81.50} & \textbf{0.10} & 0.00 & 0.00 & 0.00\\
\cline{2-12}
 & \emph{Class=1} & 75.53 & 0.00 & 0.76 & 1.00 & 0.86 & \textbf{84.00} & \textbf{0.20} & \textbf{0.84} & 1.00 & \textbf{0.91}\\
\cline{2-12}
 & \emph{Average} & 75.27 & 0.00 & 0.38 & 0.50 & 0.43 & \textbf{82.75} & \textbf{0.15} & \textbf{0.42} & 0.50 & \textbf{0.46}\\
\hline
\multirow{3}{*}{\textbf{OneR}} & \emph{Class=0} & 74.33 & 0.09 & 0.33 & 0.12 & 0.16 & \textbf{77.25} & 0.03 & 0.00 & 0.00 & 0.00\\
\cline{2-12}
 & \emph{Class=1} & 74.32 & 0.07 & 0.77 & 0.95 & 0.85 & \textbf{80.75} & \textbf{0.13} & \textbf{0.84} & \textbf{0.96} & \textbf{0.89}\\
\cline{2-12}
 & \emph{Average} & 74.33 & 0.08 & 0.55 & 0.54 & 0.51 & \textbf{79.00} & 0.08 & 0.42 & 0.48 & 0.45\\
\hline
\end{tabular}
}
\label{tab:results2}
\end{center}
\end{table}

\begin{figure}
  \includegraphics[width=3.5in]{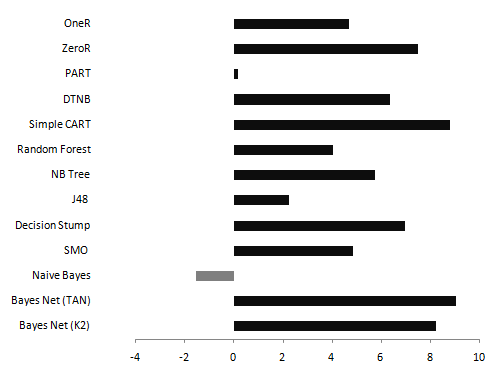}
  \caption{Delta between average CCI\% Aljezur and Aljezur SPPAM }
  \label{fig:aljezur_delta}
\end{figure}

In Figure~\ref{fig:aljezur_delta} we show the difference between the
averages of correctly classified instances percentage for the
original dataset and the SPPAM-preprocessed dataset for the Aljezur
dataset. Here we can see graphically how better in average, the
classification algorithms can correctly classify new instances using
our method. The values are also in percentage.

\section{Conclusions and Future Work}
\label{conc}

In this work, we proposed a simple, general solution to the problem of
learning classifiers for multiple correlated data such as multiple
exams of a single patient, internet customer preferences, weather
forecast, sea prediction, among others. SPPAM, a Statistical
PreProcessing AlgorithM, takes the original dataset containing related
data, and produces a new dataset with all correlated data aggregated
using metrics such as maximum, minimum, average, etc. We tested SPPAM
on two datasets that contain records associated according to a
date. We used WEKA to train on the original datasets and on the
SPPAM-transformed dataset. Our results indicate that the SPPAM
transformation can produce better classifiers than the ones trained on
the original dataset.

In its present form, SPPAM has already shown its potential, but we
have been working on modifications to the basic algorithm in order to
improve performance even further. We also have been working on
applying SPPAM to medical datasets that contain multiple records for
a single patient.

{\footnotesize }

{\footnotesize
\bibliographystyle{plain}
\bibliography{sppam}
}

\end{document}